%% file: main.tex
\documentclass[10pt]{article}
\usepackage{preprint}

\usepackage{siunitx}

\input{math_commands}

\graphicspath{{./}}

\title{CascadeFormer: Depth-Tapered Transformers Motivated by
Gradient Fan-in Asymmetry}
\author{%
\textbf{Huzama Ahmad}\correspondence{huzama@huzama.com} \quad
\textbf{Cao Viet Hai Nam} \quad
\textbf{Se-Young Yun} \\[3pt]
\normalsize KAIST%
}

\begin{document}
\maketitle

\begin{abstract}
\input{abstract}
\end{abstract}

\input{sections/00_introduction}
\input{sections/01_related_work}
\input{sections/02_gfa}
\input{sections/03_experiments}

\input{sections/04_discussion}
\input{sections/05_limitations}

\bibliography{references,extra,pruning,custom}

\appendix
\input{appendix/00_llm_usage}
\input{reproducibility}
\input{appendix/01_experiment_details}
\input{appendix/02_additional_results}

\end{document}

%% file: math_commands.tex
\usepackage{amsmath,amsfonts,bm}

\def\eqref#1{equation~\ref{#1}}

\def\1{\bm{1}}

\DeclareMathAlphabet{\mathsfit}{\encodingdefault}{\sfdefault}{m}{sl}
\SetMathAlphabet{\mathsfit}{bold}{\encodingdefault}{\sfdefault}{bx}{n}



%% file: abstract.tex
Deep Transformers are composed of uniformly stacked residual blocks, yet their deepest layers often add little value. We present two efficiency methods that exploit this asymmetry. CascadeFormer tapers width with depth to match the uneven information flow across layers, achieving comparable perplexity to a uniform baseline at the same training budget while reducing latency by 8.6\% and increasing throughput by 9.4\%. CascadeFlow Pruning removes layers using accumulated training gradients, with no post hoc analysis. It outperforms standard heuristics on perplexity and rank-stability and stays competitive on downstream accuracy. To motivate these methods, we propose \textit{Gradient Fan-in Asymmetry} (GFA) as a structural account of why deeper layers contribute less. In Pre-LayerNorm residual stacks, the gradient at a layer is the sum of an identity path and all downstream functional paths, producing a gradient fan-in that decays linearly with depth (and quadratically under deep supervision), yielding richer gradients for early layers and sparser ones for later layers. We provide correlational and interventional evidence for GFA on models trained from scratch up to 1.2B parameters. Across Transformers and ResNets, accumulated training gradients follow the theoretical fan-in and are associated with post hoc layer importance. Two interventions point to structure rather than magnitude as the bottleneck: equalizing per-layer gradient norms does not restore late-layer value, while increasing downstream path counts via parameter-shared repetition restores and elevates it. Whether gradient magnitude proxies fan-in beyond high-rank regimes, and how these dynamics behave at the 100B+ scale, remain open questions.

%% file: sections/00_introduction.tex
\section{Introduction} \label{sec:intro}

\begin{figure}[htbp]
\centering
\includegraphics[width=0.9\textwidth]{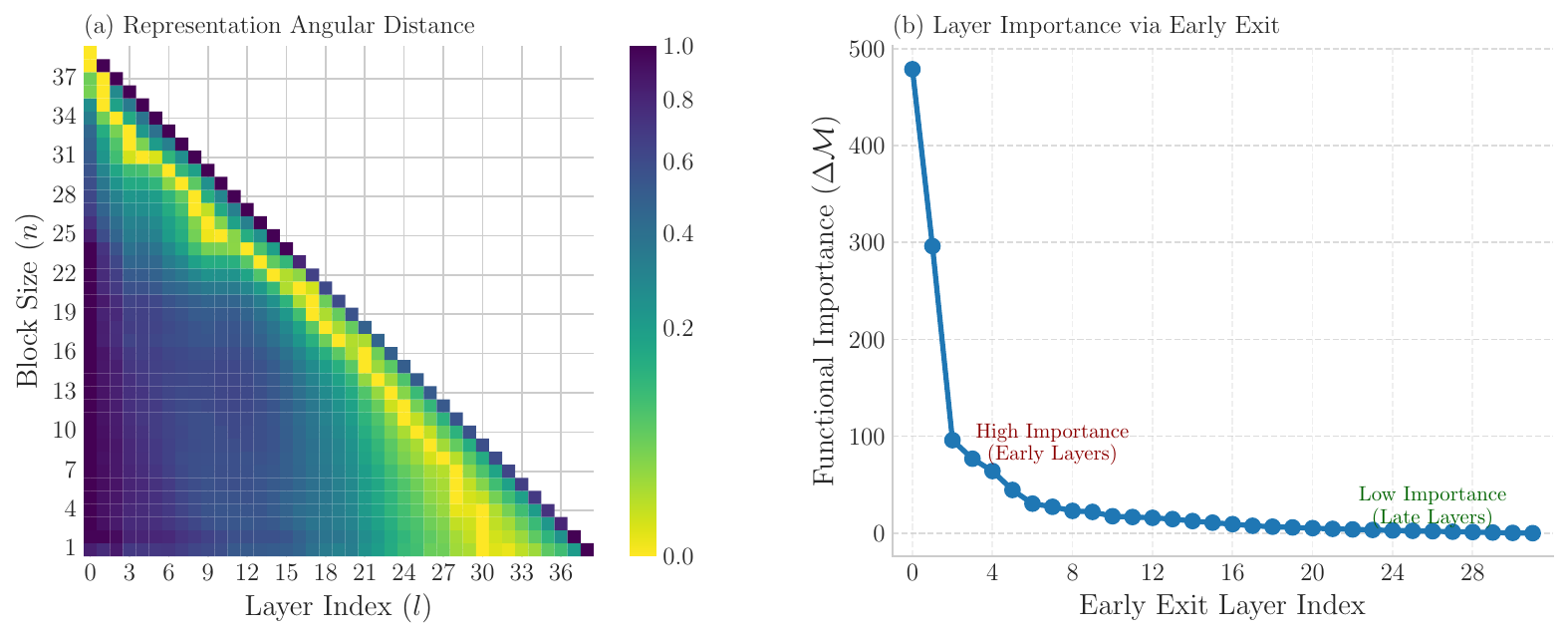}
\caption{\textbf{Deeper Transformer layers show diminishing contributions.} Illustrative motivation from pretrained public checkpoints (distinct from the from-scratch models studied in the rest of the paper). (a) In a pretrained LLaMA 13B model, representational similarity across layers increases with depth, signalling growing redundancy. (b) A pretrained LayerSkip LLaMA 8B (an early-exit/deep-supervised model, where the effect is pronounced by design) makes the consequence explicit: layer importance, measured by Functional Importance ($\Delta\mathcal{M}$) upon its removal, is concentrated in the initial layers while the functional value of later layers collapses.}
\label{fig:intro_layer_importance}
\end{figure}

The uniform scaling of transformer blocks \cite{vaswani_attention_2017}, simply repeating identical layers to create deeper models, has been the driving principle behind the success of Large Language Models \citep{radford_language_nodate, brown_language_2020, touvron_llama_2023}. However, uniform stacking masks a functional asymmetry. For instance, when we evaluate a pretrained LLaMA model on WikiText, its deeper layers exhibit high representational similarity, a key indicator of redundancy~\citep{gromov_unreasonable_2024} (Figure~\ref{fig:intro_layer_importance}a). This asymmetry is even more pronounced in architectures like LayerSkip~\citep{elhoushi_layerskip_2024}, which skips later layers by exiting the network early, revealing their sharply declining functional contribution (Figure~\ref{fig:intro_layer_importance}b).

The conventional explanation for this phenomenon points to attenuated gradients in deeper layers \cite{li_mix-ln_2025}. The explanation is correct, but it may treat a symptom rather than the underlying factor. We propose that the relevant quantity is not gradient magnitude but the compositional diversity of the gradient, a structural bottleneck we term \textbf{Gradient Fan-in Asymmetry (GFA)}. Residual connections \citep{he_deep_2016} turn the network into an implicit ensemble of paths of varying lengths \citep{veit2016residualnetworksbehavelike}, so shallow layers receive gradient from all downstream blocks while the deepest layers, aggregating from few, receive a structurally simple, information-poor gradient.

We support the GFA hypothesis and demonstrate its utility through a sequence of empirical arguments, all on models trained from scratch up to 1.2B parameters. First, we observe a positive correlation between per-layer gradient norms $\bar{g}_i$ and eventual functional importance $\Delta\mathcal{M}_i$. We then move beyond correlation with two interventional tests: one ablative, showing that artificially amplifying late-layer gradient magnitude fails to restore late-layer importance, and one constructive, showing that structurally increasing their path counts via layer repetition does restore it. These interventions are consistent with a structural (not magnitude) bottleneck. Finally, we translate the GFA account into two practical applications: \textbf{CascadeFlow Pruning (CFP)}, an efficient method leveraging accumulated training gradients to outperform standard pruning heuristics, and the \textbf{CascadeFormer}, an architecture that tapers width with depth to align model capacity with the natural flow of compositional gradient diversity, improving inference efficiency at fixed training FLOPs.

We reframe gradient magnitude as a proxy for a structural information imbalance, not as a cause to be fixed. Our contributions are:
\begin{itemize}
\item \textbf{We design the CascadeFormer}, an architecture that internalizes the GFA principle. By tapering network width with depth to match the natural flow of compositional gradient diversity, it reduces latency and increases throughput over a uniform baseline with equal training FLOPs and comparable perplexity.

\item \textbf{We introduce CascadeFlow Pruning (CFP)}, an efficient method that leverages accumulated training gradients as a proxy for structural importance to prune layers, outperforming standard heuristics without requiring expensive post hoc analysis.

\item \textbf{We propose Gradient Fan-in Asymmetry as a structural account} of layer redundancy and provide correlational and interventional evidence for it on models trained from scratch up to 1.2B parameters. Two complementary interventions, an ablative test (artificially equalizing gradient norms fails to restore importance) and a constructive one (structurally increasing path counts via layer repetition succeeds), are consistent with a bottleneck rooted in the gradient's compositional complexity rather than its raw magnitude. Whether magnitude proxies fan-in beyond high-rank regimes, and how these dynamics behave at the 100B+ scale, remain open questions.
\end{itemize}

%% file: sections/01_related_work.tex
\section{Related Work} \label{sec:literature}

\paragraph{Layer Redundancy for Model Efficiency.}
Deep networks exhibit layer redundancy that can be exploited for compression and faster inference \citep{gromov_unreasonable_2024, sun_curse_2025, chen_streamlining_2025}. Structured pruning removes entire blocks with minor loss \citep{chen_lorashear_2023, frantar_sparsegpt_2023, ma_llm-pruner_2023, xia_sheared_2024, kim_shortened_2024, he_what_2024, sun_simple_2024}, while training-time methods like LayerDrop \citep{Fan2020Reducing} and early-exit/skip mechanisms \citep{elhoushi_layerskip_2024, xin2020deebertdynamicearlyexiting, liu2020fastbertselfdistillingbertadaptive, zhao_skipgpt_2025, men2025shortgpt} allow dynamic redundancy management. These works document \emph{that} redundancy exists and how to exploit it; the mechanistic accounts that do exist attribute it largely to gradient \emph{magnitude} attenuation, whereas we propose a \emph{compositional} (fan-in) account.

\paragraph{The Architectural Roots of Redundancy.}
Residual networks \citep{he_deep_2016, he2016identitymappingsdeepresidual} can be viewed as implicit ensembles of shorter paths \citet{veit2016residualnetworksbehavelike}, a property inherited by Transformers with Pre-LayerNorm architectures \citep{xiong_layer_2020, touvron_llama_2023}. This structure often causes deeper layers to contribute minimally, leading to redundancy \citep{takase_b2t_2023, sun_curse_2025, li_mix-ln_2025}. A complementary line of work shows how architectural and normalization choices regulate gradient propagation across depth \citep{wang_deepnet_2022, shleifer_normformer_2021, li_mix-ln_2024}. Our methods target the Pre-LN setting. Building on these results, we hypothesize that the depth-wise allocation of gradient signal is structurally induced rather than driven by weight magnitude, and we provide correlational and interventional evidence for this. We then exploit the effect to design more efficient architectures along both depth and width. Observations that attention-layer redundancy is inherent and stable across training stages \citep{he_what_2024} point to a structural rather than incidental origin.

\paragraph{Quantifying Layer Importance.} Identifying redundant layers requires a reliable importance metric. Magnitude and first/second-order criteria provide strong baselines \citep{han_learning_2015, han_deep_2016, molchanov_pruning_2017, lee_snip_2019, frantar_sparsegpt_2023}. Other methods rely on measuring the effects of redundancy: output-similarity (e.g., cosine between adjacent layers), correlates high similarity with low importance \citep{gromov_unreasonable_2024, yang_laco_2024, jiang2025tracing, chen_streamlining_2025, song_sleb_nodate}, while perturbation-based metrics such as ($\Delta$PPL) upon single layer removal directly quantify its functional contribution \citep{kim_shortened_2024}. Gradient-based signals have been used as local surrogates for importance, but typically in heuristic form (e.g., saliency \cite{smilkov_smoothgrad_2017, selvaraju_grad-cam_2020}, Taylor criteria \citep{yang-etal-2023-gradient, ma_llm-pruner_2023}). We propose that gradient dynamics do more than proxy importance: through a compositional fan-in asymmetry, they may help account for the final functional hierarchy, and they predict it on the from-scratch models (up to 1.2B parameters) we study. We provide correlational and interventional evidence for this.

%% file: sections/02_gfa.tex
\section{Gradient Fan-in Asymmetry} \label{sec:GFA}

\paragraph{The Phenomenon.}
We propose Gradient Fan-in Asymmetry (GFA), a structural imbalance in the composition of gradient signals within deep residual architectures, as an account of layer redundancy. This asymmetry arises because shallow layers receive gradients aggregated across numerous downstream computational paths, while deep layers receive them from a progressively smaller set. We hypothesize that this disparity in fan-in shapes the \textit{compositional diversity} of the resulting gradient. Under this account, deep-layer gradients are compositionally simpler and thus information-poorer, leading to less effective weight updates. The limitation is structural, not one of magnitude. An optimizer that only rescales gradients cannot supply the missing compositional information.

\paragraph{GFA in Residual Networks.}
We define a Pre-LN Transformer as $x_{l+1} = x_l + F_l(x_l)$, where $F_l$ represents the block's complete transformation, including LayerNorm and sublayers. The gradient at its input, $g_l \equiv \partial \mathcal{L}/\partial x_l$, unrolls into a cumulative sum over all downstream blocks:
\begin{equation}
g_l \;=\; g_N \;+\; \sum_{k=l}^{N-1} {J_k^T} g_{k+1},
\label{eq:gradient_sum}
\end{equation}
where $J_k$ is the Jacobian of the $k$-th block's transformation. This structure means the gradient at layer $l$ aggregates signals from an identity path and all subsequent functional paths. We term the number of these aggregated signals the gradient fan-in, $\phi_l$, which decreases linearly with depth (visualized by the solid paths in Fig.~\ref{fig:unwrapped_transformer}). A formal counting rule is provided in Appendix~\ref{app:pathcounts}.

\paragraph{Amplification via Deep Supervision.}
Architectures employing deep supervision, such as LayerSkip, amplify GFA. By introducing auxiliary loss heads at intermediate layers, they create new gradient hierarchies that backpropagate to shallower layers. This transforms the linear fan-in disparity into a quadratic one (illustrated by the additional dotted paths in Fig.~\ref{fig:unwrapped_transformer}), concentrating gradient information in the shallowest layers. Appendix~\ref{app:pathcounts} gives the full derivation. Deep supervision therefore exacerbates the structural imbalance rather than solving it.

\paragraph{Analysis and Prediction.}
Our fan-in proxy is not the $2^{N-l}$ combinatorial paths of a full expansion. Our proxy measures the number of distinct signal channels aggregated at a layer, not their information quality (e.g., orthogonality), which remains an empirical question. The distinction motivates our central hypothesis: \textit{gradient norm may be a symptom rather than the underlying factor}. Under this account, a small gradient norm in a deep layer reflects its limited access to compositional information. An optimizer like AdamW \cite{loshchilov_decoupled_2018} can rescale this gradient, but it cannot invent compositional information that is structurally absent. From this account, we derive a testable prediction: an architectural change that structurally increases a layer's downstream gradient fan-in should increase its functional importance. We test this prediction directly with virtual depth and use the account to motivate our methods.

\begin{figure}[t!]
\centering
\includegraphics[width=0.90\linewidth]{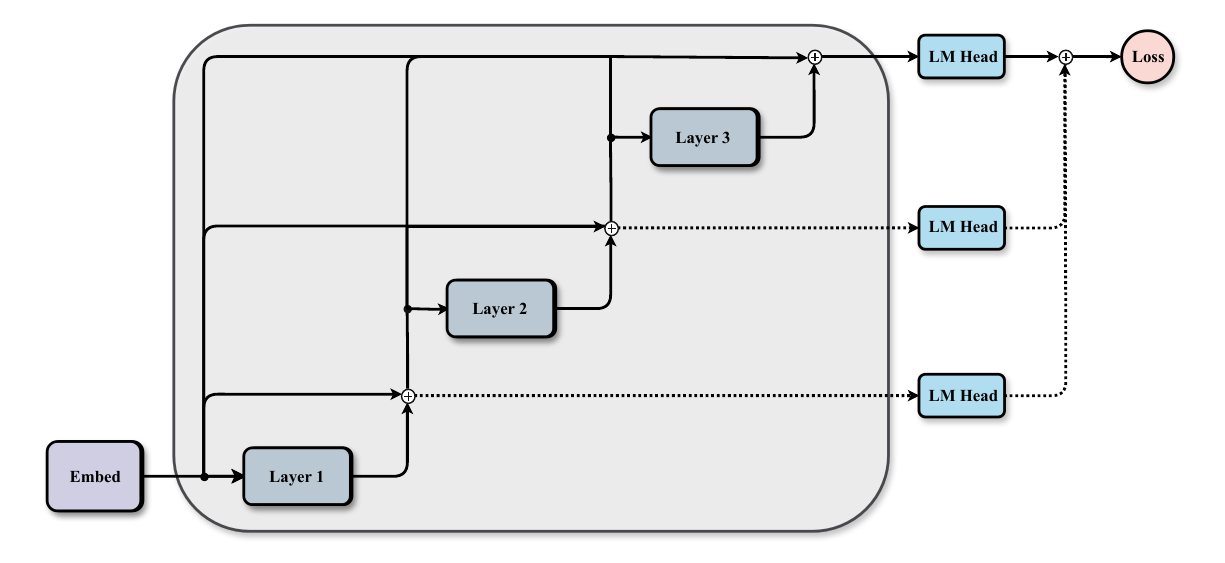}
\caption{\textbf{Gradient Fan-in Asymmetry arises from a structural imbalance in gradient paths.} The unwrapped view shows that the gradient at any layer $l$ is a sum over signals from an identity path and all subsequent functional paths (Eq.~\ref{eq:gradient_sum}). The number of contributing functional paths (solid lines) decreases linearly with depth, creating a compositional asymmetry. Deep supervision (dotted lines) exacerbates this imbalance by introducing new gradient hierarchies from auxiliary losses, which transforms the fan-in disparity from linear to quadratic.}
\label{fig:unwrapped_transformer}
\end{figure}

%% file: sections/03_experiments.tex
\section{Empirical Evidence and Applications} \label{sec:experiments_and_results}

We test Gradient Fan-in Asymmetry in three stages. We first show that gradient flow is structurally skewed towards early layers and that this training signal predicts the final functional hierarchy. We then perform two interventions that separate magnitude from information content and are consistent with structure as the bottleneck. Finally, we translate the GFA account into two applications, CascadeFlow Pruning and CascadeFormer, that improve efficiency at fixed training cost.

\subsection{Setup} \label{ssec:setup}

\paragraph{Models and architectures}
We evaluate three residual families. For language modelling we train a sixteen-layer, approximately 1.2B-parameter, Llama base Transformer \citep{touvron_llama_2023} referred to as Vanilla and a LayerSkip variant \citep{elhoushi_layerskip_2024}. For vision we train a ResNet-50 \citep{he_deep_2016}. To encode the GFA prior we modify the Vanilla architecture to create CascadeFormer which tapers width with depth to align capacity with the decay in compositional gradient diversity. All models are trained from scratch.

\paragraph{Datasets and tasks}
Language models are trained on a seven billion token subset of Dolma \citep{soldaini_dolma_2024} for next token prediction. ResNet-50 is trained on ImageNet-1K \citep{deng_imagenet_2009}. Training hyperparameters and optimizer settings are in Appendix~\ref{appx:experiment_details}. For our primary architectural comparison, we train the proposed CascadeFormer\textsubscript{A\textsubscript2}, the full baseline, and a 15-layer Baseline, using three different random seeds.

\subsection{Quantifying Gradient Flow and Functional Importance} \label{ssec:metrics_formal}

To empirically test our hypothesis, we require metrics that can directly link training dynamics with the final functional hierarchy of the model.

\paragraph{Accumulated Gradient Share ($\bar{g}_i$).} To capture a layer's overall contribution during training, we accumulate the L2 norm of the gradients with respect to its parameters, $\theta_i$, over $T$ training steps. This accumulated value is then normalized by the total sum from all $N$ layers to yield the relative gradient share, $\bar{g}_i$:
\begin{equation}
\bar{g}_i = \frac{\sum_{t=1}^{T} \lVert \nabla_{\theta_i}\mathcal{L}_t \rVert_2}{\sum_{j=1}^{N} \sum_{t=1}^{T} \lVert \nabla_{\theta_j}\mathcal{L}_t \rVert_2}.
\label{eq:grad_share}
\end{equation}
This metric serves as a direct, data-driven proxy for the structural information flow predicted by GFA.

\paragraph{Functional Importance ($\Delta\mathcal{M}_i$).} We quantify a layer's functional importance by measuring the performance degradation when its contribution is removed. This is achieved by ablating layer $i$, bypassing its computational block while preserving the residual path to the subsequent layer. The functional importance, $\Delta\mathcal{M}_i$ is the absolute degradation in the task metric $M$ resulting from this ablation. For language models, this is the increase in perplexity ($\Delta$PPL), and for vision models, the change in top-1 accuracy ($\Delta$Acc). A larger $\Delta\mathcal{M}_i$ signifies greater functional importance.

\begin{figure}[t]
\centering
\includegraphics[width=0.8\textwidth]{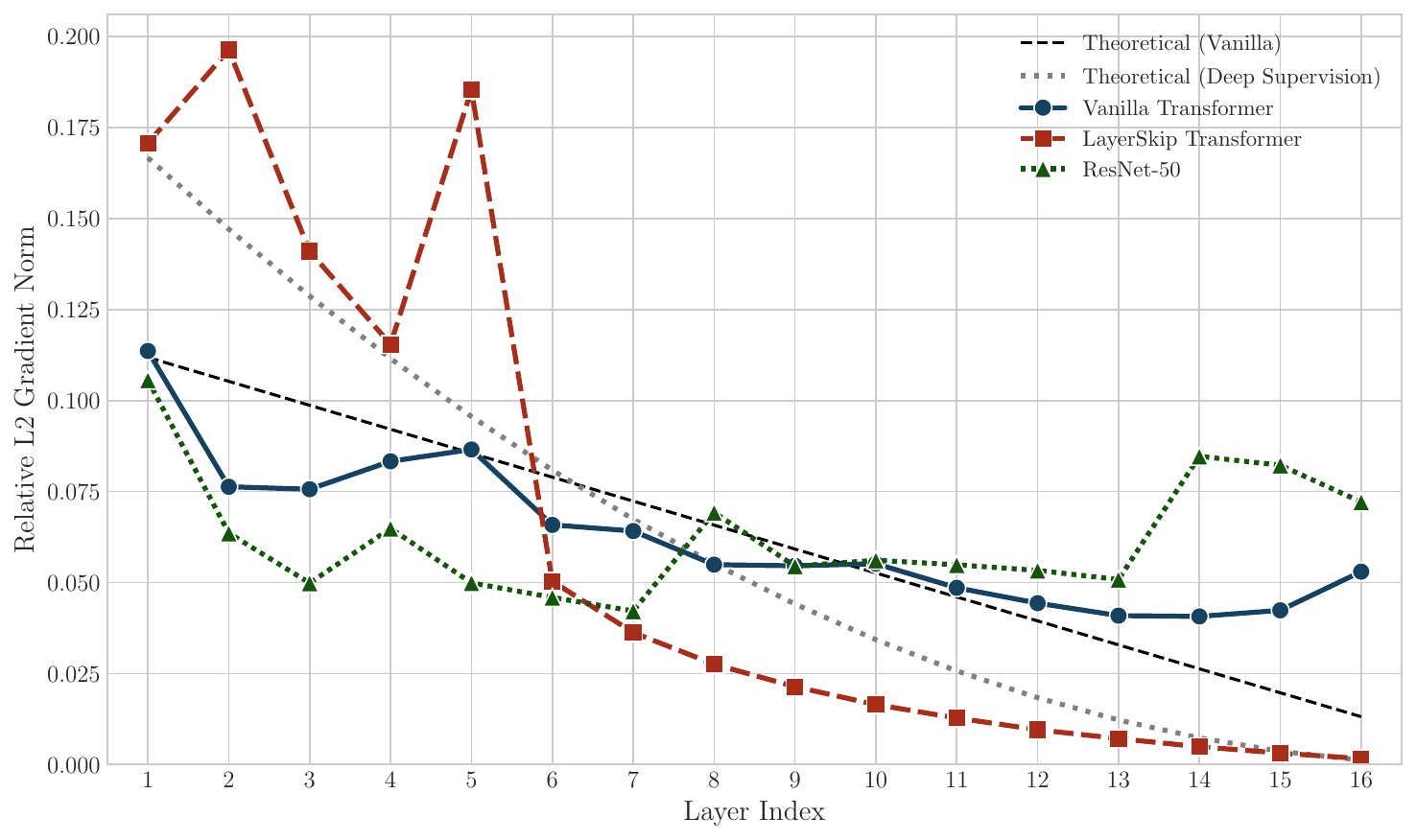}
\caption{\textbf{Gradient flow is inherently front-loaded in deep residual architectures.} The average L2 gradient norm per layer in a Vanilla Transformer follows a linear decay, closely tracking its theoretical fan-in (black, dashed). In contrast, LayerSkip's deep supervision mechanism induces a quadratic decay, a behaviour accurately modelled by its own theoretical curve (grey, dotted). ResNet-50 shows the same front-loaded decay, tracking the structural fan-in predicted by GFA.}
\label{fig:gradient_flow}
\end{figure}

\subsection{Correlational evidence across architectures} \label{ssec:correlation}

\paragraph{Structural fan-in aligns with gradient flow}
GFA predicts that the downstream gradient fan-in decreases with depth, inducing a front-loaded gradient distribution. Figure~\ref{fig:gradient_flow} illustrates this empirical pattern across three architectures alongside their theoretical fan-in decay. For the vanilla network, this decay is linear, while for LayerSkip it is quadratic. We observe that the theoretical fan-in and the empirical gradient norm $\bar{g}_i$ share the same monotonic ordering. Complete derivations and path counts are provided in Appendix~\ref{app:pathcounts}.

\begin{figure}[t]
\centering
\includegraphics[width=\textwidth]{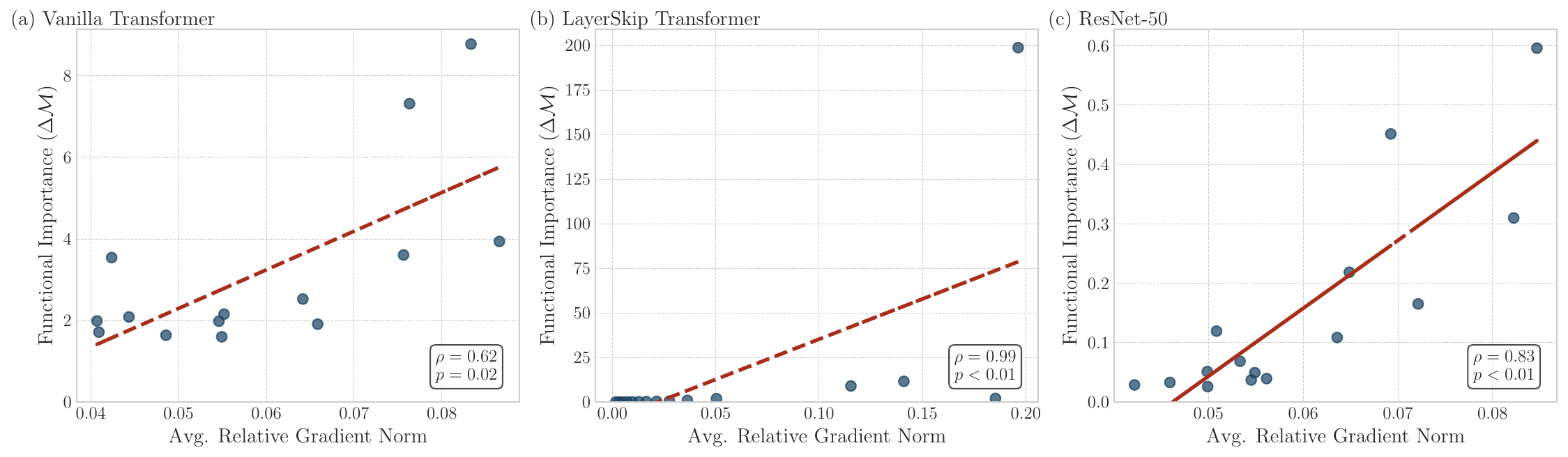}
\caption{\textbf{Training gradient flow predicts final layer importance.} The accumulated gradient share $\bar{g}_i$ during training shows a strong Spearman correlation with post hoc functional importance $\Delta\mathcal{M}_i$. The correlation is significant in the Vanilla Transformer with $\rho = 0.62$ and in ResNet-50 with $\rho = 0.83$, and is near perfect in LayerSkip with $\rho = 0.99$.}
\label{fig:grad_importance_correlation}
\end{figure}

\paragraph{Gradient flow is associated with the functional hierarchy}
We correlate each layer’s $\bar{g}_i$ with its ablation-based importance $\Delta\mathcal{M}_i$. Figure~\ref{fig:grad_importance_correlation} shows a strong positive Spearman correlation for the Vanilla Transformer with $\rho = 0.62$ and $p = 0.02$ and for ResNet-50 with $\rho = 0.83$ and $p < 0.01$. The relationship tightens in LayerSkip with $\rho = 0.99$ and $p < 0.01$ where supervision accentuates early-layer fan-in. Because LayerSkip's auxiliary losses attach directly to early layers, this near-perfect correlation partly reflects that design and is best read as a consistency check rather than independent confirmation; the Vanilla ($\rho{=}0.62$) and ResNet-50 ($\rho{=}0.83$) results, which lack this mechanism, are the load-bearing correlational evidence. This links the structural gradient skew during training to the final functional hierarchy, and suggests the accumulated gradient share as a predictive proxy for functional importance on the models studied here.

\subsection{Interventions separate magnitude from structure} \label{ssec:interventions}

Correlation does not imply causation. To disentangle the roles of gradient magnitude and structure, we conduct two interventional tests designed to probe the GFA hypothesis.

\paragraph{Equalizing magnitude does not restore importance}

\begin{figure}[t]
\centering
\includegraphics[width=\textwidth]{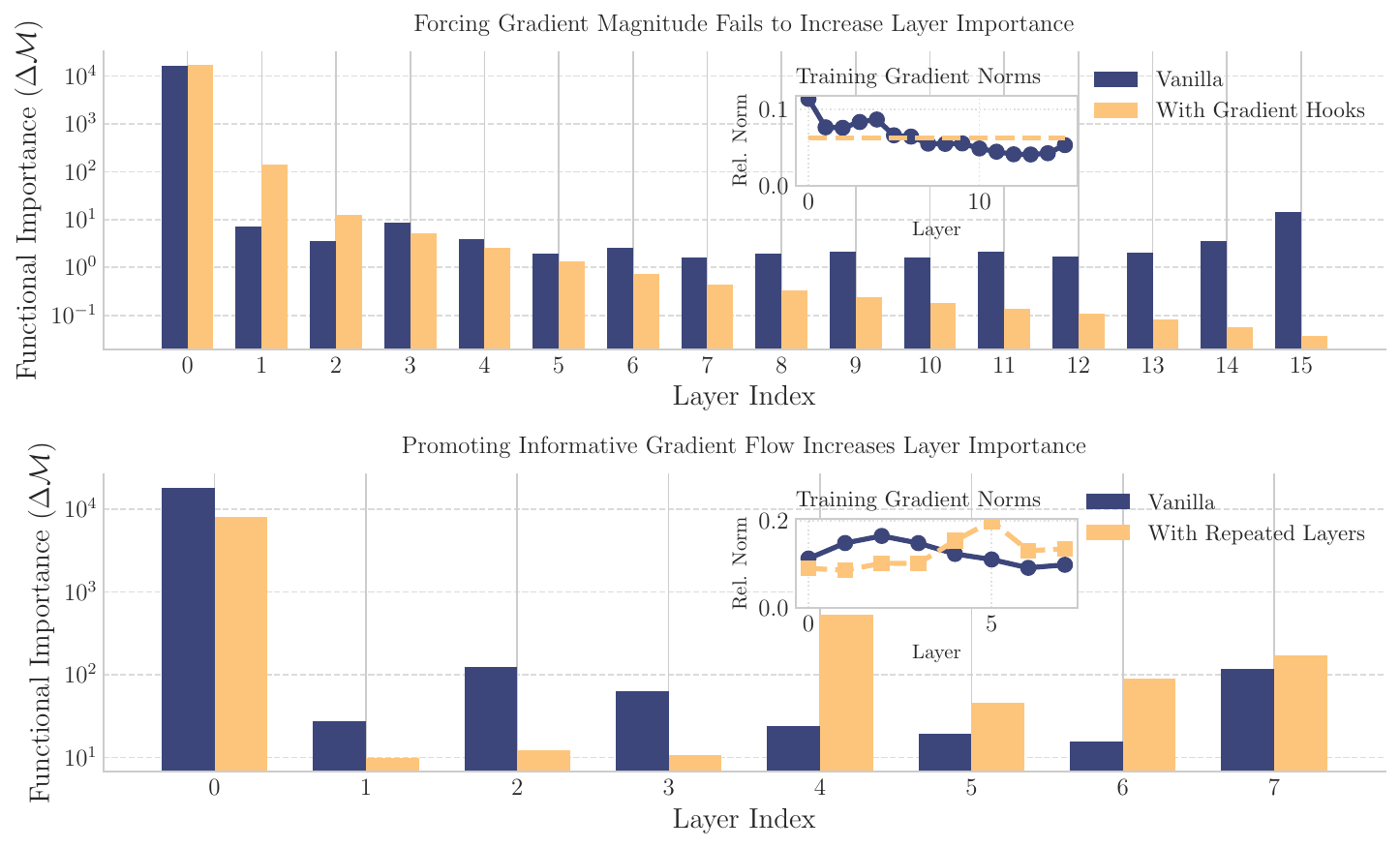}
\caption{\textbf{Interventional tests are consistent with gradient structure shaping importance.} Top equalizing L2 gradient norms across layers does not rescue and reduces deep layer importance. Bottom increasing downstream path counts by parameter shared repetition restores and elevates late layer importance. A logarithmic y-axis spans the wide dynamic range. What matters is how each layer's importance changes relative to its baseline.}
\label{fig:hook_ablation_16}
\end{figure}

We first test the alternative hypothesis: that small gradient magnitude is the direct cause of redundancy. During training, we insert a hook that rescales per-layer gradients to have an equal L2 norm. To accommodate this artificial amplification, we proportionally scaled the gradient clip norm, deriving the factor from the ratio of maximum observed norms between the hooked and standard models, details shown in Appendix~\ref{app:causal_intervention}. If magnitude were the causal factor, this should rescue the importance of later layers. The result, shown in Figure~\ref{fig:hook_ablation_16} (top), is the opposite. The figure shows a logarithmic comparison against a vanilla model, with an inset detailing the validation gradient distribution; equalizing the norms not only fails to restore importance but actively harms the contribution of deep layers. Amplifying an information-poor signal does not add compositional content; it only increases the norm and can destabilize learning.

\paragraph{Increasing path counts restores importance}
Next, we directly test the structural component of GFA against a vanilla 8-layer reference. We engineer an increase in the downstream gradient paths for deep layers by repeating the final four layers of an 8-layer model with shared parameters, which increases a layer's virtual depth and its gradient fan-in without adding parameters. Specifically, we repeat the last four layers with a pattern of two, three, three, and five repeats respectively, a pattern designed to increase the fan-in with depth. This modification dramatically alters the structural fan-in for the deepest layers (detailed in Appendix~\ref{app:pathcounts}), for instance changing the counts for layers 5 through 8 from [5, 4, 3, 2] to [27, 33, 24, 20]. GFA predicts these layers should become more functionally important.

Figure~\ref{fig:hook_ablation_16} (bottom) supports this prediction. The inset shows that gradients in the repeated deep layers increase substantially, even exceeding those of early layers. Consequently, their functional importance rises, diminishing the relative contribution of the initial four layers. These modified deep layers, now recipients of a more compositionally complex gradient, become more critical than the untouched shallow layers. Together, these interventions provide interventional evidence consistent with functional hierarchy being shaped by the structural flow of gradient information, for which magnitude is a correlated proxy.

\subsection{Applications informed by GFA} \label{ssec:applications}

We now apply GFA to two efficiency methods: a pruning rule and a tapered architecture.

\paragraph{CascadeFlow Pruning uses training dynamics}
We prune layers using the accumulated gradient share, $\bar{g}_i$, gathered directly during training. Our method, CascadeFlow Pruning (CFP), uses the L2 norm of these gradients (Eq.~\ref{eq:grad_share}) as a proxy for a layer's functional importance, ranking them accordingly. This approach eliminates the need for expensive post hoc computation required by alternative heuristics like hidden state similarity \cite{gromov_unreasonable_2025}, Taylor based methods \cite{ma_llm-pruner_2023}, or parameter magnitude pruning \cite{han_learning_2015}. CFP then removes the lowest ranked layers, irrespective of their original consecutive block structure.

\subparagraph{Evaluation}
We evaluated pruning strategies on the Dolma 2.6M token holdout set and evaluation set of HellaSwag benchmark. We report perplexity (PPL) on this Dolma 2.6M holdout, on which the unpruned baseline scores 17.94. On HellaSwag, we adopt the zero-shot protocol from OLMo~\cite{olmo20242olmo2furious}, ranking candidates by their relative conditional log-probability. While this zero-shot approach does not measure post-fine-tuning adaptability, it provides a direct and efficient benchmark for quantifying performance degradation as a function of pruning.

Table~\ref{tab:pruning} summarizes the results. The similarity method becomes competitive only under the most aggressive pruning, but typically at a higher perplexity cost. In contrast, Taylor and Magnitude methods are unstable. Their reliance on a per-layer forward pass yields importance rankings that fluctuate across seeds, leading to misrankings and sharp performance degradation. Our CFP avoids this sensitivity, producing stable rankings and a more graceful degradation profile.

\subparagraph{Implementation via Layer Passthrough}
Pruning a Transformer layer in our framework requires no architectural modification: we implement it by a simple \emph{passthrough} in the forward pass since skipped layers are simply treated as identity functions. Specifically, during inference the model iterates over the stack of decoder layers. For each index $i$, if $i$ belongs to the pruned set $\mathcal{I}_{\text{prune}}$, we skip the forward call to the corresponding block and directly forward the input hidden state to the next layer:
\[
h_{i+1} \;=\; h_{i}, \qquad \text{if } i \in \mathcal{I}_{\text{prune}},
\]
This design choice allows CFP to be integrated into existing Transformer codebases (e.g., \texttt{LLaMA}-style decoders) with only a few lines of modification.

\paragraph{CascadeFormer.}
Our second application internalizes the GFA principle directly into the model's architecture. Because compositional gradient diversity decays with depth, uniform capacity allocation is inefficient. We therefore designed the \textbf{CascadeFormer}, an architecture that tapers model width to align its capacity with this information flow. For a model with $N$ layers, indexed $l \in \{0, \dots, N-1\}$, we apply tapering rules to either the attention, FFN sublayers or both.

\subparagraph{Attention Tapering.} We reduce the number of attention heads, and thus the attention dimension $d_{\text{attn}}(l)$, in discrete steps governed by,
$
d_{\text{attn}}(l) = d_{\text{attn},0} - S_d \cdot \lfloor l / F_d \rfloor,
\label{eq:attn_taper}
$
where $d_{\text{attn},0}$ is the initial dimension, $S_d$ is the dimensional reduction per step, and $F_d$ controls the frequency of the reduction.

\subparagraph{FFN Tapering.} We reduce the FFN's inner dimension $d_{\text{ffn}}(l)$ linearly with depth according to the rule,
$
d_{\text{ffn}}(l) = d_{\text{ffn},0} - S_f \cdot l,
\label{eq:ffn_taper}
$
where $d_{\text{ffn},0}$ is the initial dimension and $S_f$ controls the steepness of the linear taper.

\input{tables/tab_pruning}

\subparagraph{Variant Configurations.}
We define six CascadeFormer variants based on these rules, categorized by low (subscript 1) and high (subscript 2) tapering intensity. The specific hyperparameters for each variant are detailed in Table~\ref{tab:tapered_configs}. The combined variants (C) apply both attention and FFN tapering schemes simultaneously. To ensure a fair comparison, we also trained six baseline models whose computational cost was scaled linearly by reducing their layer count.

Table~\ref{tab:CascadeFormer_results} quantifies the performance and efficiency of our GFA-informed architectures. Our CascadeFormer-$A_2$ was designed with a training FLOP budget equivalent to the Vanilla-15L baseline. It achieves comparable perplexity to the baseline ($17.84\pm0.02$ vs $17.84\pm0.03$) while reducing inference latency by 8.6\% and increasing throughput by 9.4\%. These efficiency figures are specific to a single A100 with \texttt{torch.compile}-selected kernels and to the Vanilla-15L FLOP-matching; absolute gains may vary with hardware, batch size, and baseline choice.

To further understand the design space, we explored additional variants applying the tapering principle to FFN layers (F\textsubscript{1}, F\textsubscript{2}) and in combination (C\textsubscript{1}, C\textsubscript{2}). While all GFA-informed models are competitive, the attention-tapered variants (A\textsubscript{1}, A\textsubscript{2}) perform best, which is consistent with the self-attention mechanism being the most effective target for tapering; we note, however, that head-count reduction also yields the largest kernel-efficiency gains, so our experiments do not isolate the two effects.

\paragraph{Latency Measurement Protocol.} All inference metrics were measured on a single A100 GPU. To ensure fair comparison, we used a consistent batch size, context length, and generation length for all models. We leveraged \texttt{torch.compile} with the \texttt{mode='max-autotune'} and \texttt{fullgraph=True} options to minimize implementation-specific overhead and accurately reflect the inherent hardware-friendliness of each architecture. We report the median of 100 timed runs after a warmup phase of 10 steps to measure accurate execution time.

\input{tables/tab_cascadeformer_results}

%% file: tables/tab_pruning.tex
\begin{table}[t]
\centering
\small
\sisetup{
  mode=text,
  separate-uncertainty,
  table-align-uncertainty=true,
  detect-weight=true, detect-family=true
}
\newcommand{\best}[1]{\bfseries #1}
\setlength{\tabcolsep}{4pt}
\begin{tabular}{
    c
    c
    S[table-format=3.3(5)]
    S[table-format=3.3(5)]
    S[table-format=4.4(8)]
    S[table-format=4.4(8)]
    @{}
}
\toprule
\textbf{$k$} & \textbf{Metric} & {\textbf{CFP (Ours)}} & {\textbf{Sim}} & {\textbf{Taylor}} & {\textbf{Magnitude}} \\
\midrule

\multicolumn{2}{@{}l}{\textit{Baseline (k=0)}} & \multicolumn{4}{l@{}}{PPL=17.94 $\pm$ 0.00, Acc.=0.39 $\pm$ 0.00} \\
\midrule

\multirow{2}{*}{1}
& PPL $\downarrow$ & \best{19.848 \pm 0.082} & 21.945 \pm 0.462 & 24.700 \pm 0.559 & 24.361 \pm 2.357 \\
& Acc. $\uparrow$  & 0.381 \pm 0.002 & 0.381 \pm 0.002 & \best{0.384 \pm 0.001} & 0.362 \pm 0.013 \\
\cmidrule(l){2-6}

\multirow{2}{*}{2}
& PPL $\downarrow$ & \best{23.226 \pm 0.106} & 28.480 \pm 0.331 & 127.744 \pm 15.148 & 41.875 \pm 4.822 \\
& Acc. $\uparrow$  & \best{0.372 \pm 0.001} & 0.366 \pm 0.001 & 0.369 \pm 0.003 & 0.325 \pm 0.006 \\
\cmidrule(l){2-6}

\multirow{2}{*}{4}
& PPL $\downarrow$ & \best{59.790 \pm 1.840} & \best{59.790 \pm 1.840} & 4715.072 \pm 4166.454 & 332.174 \pm 216.582 \\
& Acc. $\uparrow$  & 0.334 \pm 0.004 & 0.334 \pm 0.004 & \best{0.336 \pm 0.002} & 0.285 \pm 0.010 \\
\cmidrule(l){2-6}

\multirow{2}{*}{6}
& PPL $\downarrow$ & \best{167.006 \pm 9.205} & 180.862 \pm 27.304 & 1193.159 \pm 547.709 & 3099.530 \pm 1708.037 \\
& Acc. $\uparrow$  & 0.299 \pm 0.004 & 0.304 \pm 0.003 & \best{0.305 \pm 0.002} & 0.269 \pm 0.002 \\
\cmidrule(l){2-6}

\multirow{2}{*}{8}
& PPL $\downarrow$ & \best{911.748 \pm 55.001} & \best{911.748 \pm 55.001} & 1403.434 \pm 434.894 & 1237.212 \pm 132.241 \\
& Acc. $\uparrow$  & \best{0.285 \pm 0.001} & \best{0.285 \pm 0.001} & 0.278 \pm 0.006 & 0.264 \pm 0.001 \\
\bottomrule
\end{tabular}
\caption{\textbf{CFP achieves the lowest perplexity and the most stable rankings under aggressive layer pruning.} We evaluate CFP (Ours) against standard pruning heuristics by removing an increasing number of layers ($k$). CFP consistently achieves the lowest perplexity (PPL) and maintains competitive downstream accuracy (Acc), particularly at higher sparsities.}
\label{tab:pruning}
\end{table}

%% file: tables/tab_cascadeformer_results.tex
\begin{table}[ht]
    \centering
    \small
    \setlength{\tabcolsep}{4pt}
    \begin{tabular}{l cc cccc}
        \toprule
        & \multicolumn{2}{c}{\textbf{Core Metrics}} & \multicolumn{4}{c}{\textbf{Hardware Efficiency}} \\
        \cmidrule(lr){2-3} \cmidrule(lr){4-7}
        \textbf{Model} &
        \makecell{\textbf{PPL} \\ \small{$\downarrow$}} &
        \makecell{\textbf{Params} \\ \small{(B)}} &
        \makecell{\textbf{Util.} \\ \small{(TFLOP/s) $\uparrow$}} &
        \makecell{\textbf{Cost} \\ \small{(TFLOPs)}} &
        \makecell{\textbf{Latency} \\ \small{(ms) $\downarrow$}} &
        \makecell{\textbf{Throughput} \\ \small{(tok/s) $\uparrow$}} \\
        \midrule
        \multicolumn{7}{l}{\textit{Uniform Baselines}} \\
        \quad Vanilla-16L & 17.69 & 1.28 & 53.87 $\pm$ 0.30 & 4.82 & 89.47 $\pm$ 0.50& 22,890 $\pm$ 129 \\
        \quad \textbf{Vanilla-15L }& 17.84 & 1.21 & 54.23 $\pm$ 0.34 & 4.54 & 83.81 $\pm$ 0.53& 24,439 $\pm$ 153 \\
        \quad Vanilla-14L & 18.00 & 1.15 & 54.68 $\pm$ 0.38 & 4.27 & 78.09 $\pm$ 0.54& 26,227 $\pm$ 181 \\
        \quad Vanilla-13L & 18.17 & 1.08 & 55.18 $\pm$ 0.39 & 4.00 & 72.40 $\pm$ 0.52& 28,287 $\pm$ 201 \\
        \quad Vanilla-12L & 18.82 & 1.01 & 55.74 $\pm$ 0.34 & 3.72 & 66.75 $\pm$ 0.40& 30,683 $\pm$ 186 \\
        \midrule
        \multicolumn{7}{l}{\textit{CascadeFormer (Ours)}} \\
        \quad CascadeFormer-A$_1$ & 17.79 & 1.25 & 55.39 $\pm$ 0.37& 4.72 & 85.17 $\pm$ 0.57& 24,048 $\pm$ 161 \\
        \quad \textbf{CascadeFormer-A$_2$} & 17.84 & 1.22 & 59.77 $\pm$ 0.38& 4.58 & 76.62 $\pm$ 0.48& 26,731 $\pm$ 169 \\
        \addlinespace
        \quad CascadeFormer-F$_1$ & 17.88 & 1.19 & 51.16 $\pm$ 0.27& 4.43 & 86.65 $\pm$ 0.47& 23,635 $\pm$ 127 \\
        \quad CascadeFormer-F$_2$ & 18.10 & 1.09 & 49.92 $\pm$ 0.24& 4.05 & 81.06 $\pm$ 0.39& 25,266 $\pm$ 123 \\
        \addlinespace
        \quad CascadeFormer-C$_1$ & 17.94 & 1.16 & 53.00 $\pm$ 0.35& 4.33 & 81.71 $\pm$ 0.54& 25,066 $\pm$ 164 \\
        \quad CascadeFormer-C$_2$ & 18.30 & 1.03 & 53.88 $\pm$ 0.49& 3.81 & 70.65 $\pm$ 0.66& 28,990 $\pm$ 266 \\
        \bottomrule
    \end{tabular}
    \caption{\textbf{CascadeFormer matches the baseline's perplexity at lower latency and higher throughput.} Our CascadeFormer-$A_2$ model, which tapers its attention mechanism according to GFA principles, matches the perplexity of its Vanilla-15L baseline that was trained with an equivalent computational budget while being substantially faster, reducing inference latency by 8.6\% while increasing throughput by 9.4\%.}
    \label{tab:CascadeFormer_results}
\end{table}

%% file: sections/04_discussion.tex
\section{Discussion and Conclusion} \label{sec:conclusion}

This work reframes layer redundancy in residual networks not as a failure of optimization, but as a phenomenon that may follow from their structure. We presented evidence that Gradient Fan-in Asymmetry offers a structural account of layer redundancy, supported by correlational and interventional evidence on from-scratch models up to 1.2B parameters. The lasting contribution is a pair of efficiency methods: CascadeFlow Pruning, a more effective pruning method, and CascadeFormer, an efficient architecture that aligns its capacity with the asymmetric flow of information. The GFA account motivated both designs; a reader who is unconvinced by the account can still adopt the methods on their measured merits.

\paragraph{Design Tension and Future Directions.} Our findings expose a design tension. One path is to embrace the asymmetry, as CascadeFormer does, leading to intentionally heterogeneous architectures that allocate resources where learning dynamics can best use them. A second path is to counteract GFA, aiming to force uniform functional contribution. This second path is complicated by evidence from architectures like LayerSkip. By imposing deep supervision, these models create an extreme gradient hierarchy that forces shallow layers to become functionally self-sufficient. Shallow layers may hold latent capacity that standard end-to-end training, with its gentler GFA decay, does not fully use. The limitation may then lie in the training dynamic rather than in layer capacity. Such a pursuit would require new architectural components or training schemes that can inject compositional diversity into deep layer gradients, perhaps through novel long range information pathways or regularization techniques. Whether uniform contribution is achievable, or even desirable, remains open.

%% file: sections/05_limitations.tex
\section*{Limitations}
Our analysis frames Gradient Fan-in Asymmetry in terms of path quantity, using fan-in as a proxy for compositional diversity. This leaves two questions explicitly open. First, magnitude proxies fan-in only when the gradient is high-rank: a layer with small but high-quality (high effective-rank, near-orthogonal) gradients need not be redundant, so measuring gradient effective rank, orthogonality, and information content is the natural next step. Second, all of our evidence is on from-scratch models up to 1.2B parameters; while the GFA account is stated in architectural terms, whether it holds at the 100B+ parameter scale is an open empirical question and we do not claim universality. Our quality evidence for CascadeFormer is limited to language-model perplexity and zero-shot HellaSwag. Finally, our proposed pruning method, CFP, requires access to training-time gradients, making it inapplicable for post hoc pruning of pre-trained, closed-source models.

%% file: appendix/00_llm_usage.tex
\section*{Use of Large Language Models}
We disclose our use of Large Language Models (LLMs) in preparing this manuscript. Our use was focused on two areas: manuscript preparation and literature discovery.

\paragraph{Manuscript Preparation}
We utilized an LLM for copy-editing tasks, including correcting grammar, refining prose for clarity, and ensuring stylistic consistency. The research questions, methodology, analysis, and conclusions are the authors' own.

\paragraph{Literature Discovery}
An LLM was employed as a preliminary tool to broaden our literature search. Its function was to identify potentially relevant papers and suggest alternative search keywords based on our initial research scope. Every paper cited in this work was subsequently read, critically evaluated, and integrated into our analysis by the authors to confirm its relevance and correctness.

%% file: reproducibility.tex
\section*{Reproducibility Statement}
\label{sec:reproducibility}
To ensure the reproducibility of our findings, we document our methodology, data, and computational environment below.
\textbf{Code:} The complete codebase to reproduce all experiments, including model training, evaluation, and figure generation, will be made publicly available on GitHub upon publication. \textbf{Environment:} The hardware and software stacks for training (TPU v4) and evaluation (NVIDIA A100/A6000), including all library versions, are documented in Appendix~\ref{appx:compute_software}. \textbf{Data:} All experiments utilize public datasets. Language models were trained on the publicly available pre-tokenized Dolma dataset from Hugging Face. Vision models were trained on the standard ILSVRC 2012 ImageNet-1k dataset.
\textbf{Methodology:} Full architectural specifications for all models, including our novel CascadeFormer, are in Appendix~\ref{app:model_config} and~\ref{app:CascadeFormer_details}. The precise methodology for our gradient-equalization intervention, in which amplifying late-layer gradient magnitude does not restore late-layer importance, is detailed in Appendix~\ref{app:causal_intervention}. \textbf{Hyperparameters:} All training and evaluation hyperparameters are enumerated in Appendix~\ref{app:training_details}. All experiments can be reproduced with fixed random seed of \texttt{324709} for model initialization and data loading.

%% file: appendix/01_experiment_details.tex
\section{Experiment Details}
\label{appx:experiment_details}

We detail the experimental setup below. We detail the model architectures, training hyperparameters, and the specific implementation of our proposed CascadeFormer variants as well as layer-wise pruning heuristic details.

\subsection{Computational Resources and Software}
\label{appx:compute_software}

\subsubsection{Training Environment}
All model training was conducted on Google Cloud Platform (GCP) using a 128-core TPU v4 Pod slice. Our training stack utilized PyTorch \texttt{2.7.0} with the corresponding \texttt{torch\_xla==2.7.0} library.

To distribute training, we employed the GSPMD (\cite{xu_gspmd_2021}) partitioner. We utilized a 1D sharding strategy where model parameters, gradients, and optimizer states were fully sharded across the data-parallel dimension. This approach efficiently managed memory while scaling computation across all 128 TPU cores. All training was performed using a \texttt{float32} data type.

\subsubsection{Evaluation Environment}
All evaluations were performed on a single GPU, using either an NVIDIA A100-80GB or an NVIDIA A6000-48GB. The evaluation framework was PyTorch \texttt{2.7.0}. All experiments were conducted in \texttt{float32} precision to mitigate any possibility of precision error.

\subsubsection{Implementation Details}
Our model architecture is a modification of the Llama implementation from the Hugging Face \texttt{transformers==}\allowbreak\texttt{4.51.3} library. This same codebase served as the basis for our layer-dropping ablation studies. All experimental results were logged using the Weights \& Biases (\texttt{wandb}) platform, and figures included in this paper were generated with \texttt{matplotlib}.

\subsection{Model Configurations}
\label{app:model_config}

The core of our investigation involves three distinct architectures to test the generalizability of our claims. Their primary configurations are summarized in Table~\ref{tab:model_configs}.

\input{tables/tab_model_configs}

\subsubsection{CascadeFormer Architecture Details}
\label{app:CascadeFormer_details}

The CascadeFormer is motivated by the GFA hypothesis. While standard Transformers allocate uniform capacity to all layers, our architecture is designed under the hypothesis that compositional information flows in a cascade, reducing with depth. The CascadeFormer is designed in accordance with this hypothesized flow, tapering its width to match computational capacity to the signal's richness. This is achieved by progressively reducing the dimensions of the attention and feed-forward network (FFN) sub-layers. Based on a 16-layer Transformer baseline ($l \in \{0, 1, \dots, 15\}$), we define a set of variants with modulated tapering intensity to explore this new design space.

\input{tables/tab_tapered_configs}

\subsubsection{Dynamic Gradient Scaling Intervention}
\label{app:causal_intervention}

Our intervention mechanism operates by dynamically rescaling gradients on a layer-wise basis during training. This ensures that the gradient magnitudes are harmonized across all layers before the optimizer step. The procedure is executed as follows:

\begin{enumerate}
\item \textbf{Layer-wise Norm Computation:} For each layer $i$ in the network, we aggregate all parameter gradients associated with it. We then compute the Euclidean ($L_2$) norm of these concatenated gradients, denoted as $n_i$.

\item \textbf{Target Norm Identification:} We identify the maximum gradient norm across all layers, $n_{\text{target}} = \max_i(n_i)$. This value serves as the reference magnitude for scaling.

\item \textbf{Scaling Factor Derivation:} A scaling factor $\lambda_i$ is calculated for each layer by dividing the target norm by the layer's individual norm: $\lambda_i = n_{\text{target}} / n_i$. To handle layers with zero gradients and prevent division-by-zero, $\lambda_i$ is set to $1.0$ if $n_i=0$.

\item \textbf{In-place Gradient Application:} Finally, every parameter gradient within a given layer $i$ is multiplied in-place by its corresponding scaling factor $\lambda_i$. This operation normalizes the influence of each layer's gradient relative to the layer with the strongest signal.
\end{enumerate}

\subsection{Training Hyperparameters}
\label{app:training_details}

All models were trained using the hyperparameters detailed in Table~\ref{tab:training_hyperparams}, set per domain for competitive baselines.

\input{tables/tab_training_hyperparams}

\subsection{Pruning Heuristic Details}
\label{app:pruning_details}

To provide full transparency for the pruning comparison presented in Table~\ref{tab:pruning}, we detail each pruning heuristic below.

\paragraph{Magnitude-based Heuristic \cite{han_learning_2015}}
Classical magnitude pruning removes small-magnitude weights and is a strong baseline for sparsification. In the block-level variant used here, a block’s score is the aggregate of absolute parameter values across its trainable weights; for projection matrices, values are first reduced along the input dimension to obtain per-output scores, then aggregated within the block. Token embeddings and the LM head are excluded, and the final normalization is protected. Blocks are ranked by increasing score, and the $k$ lowest are pruned.

\paragraph{Taylor-based Heuristic \cite{ma_llm-pruner_2023}}
The first-order criterion estimates loss increase from removing a block by accumulating, on a small calibration set, the absolute elementwise product of each parameter and its gradient. Scores are summed across batches; for projection matrices, per-output reduction precedes block-level aggregation. Token embeddings are excluded and the final normalization is protected. Blocks are ranked by increasing score, and the $k$ lowest are pruned.

\paragraph{Similarity-based Heuristic. \cite{gromov_unreasonable_2025}} This heuristic is based on representational similarity analysis, computing the angular distance between adjacent layer representations. Blocks of $k$ layers whose removal yields the smallest change are considered less critical and are pruned.

\paragraph{CFP (CascadeFlow Pruning) Heuristic.} In contrast, our proposed CFP method prunes layers with the lowest accumulated L2 gradient norms from training.

%% file: tables/tab_model_configs.tex
\begin{table}[h!]
\centering
\begin{tabular}{@{}lccc@{}}
\toprule
\textbf{Parameter} & \textbf{Vanilla Transformer} & \textbf{LayerSkip Transformer} & \textbf{ResNet-50} \\ \midrule
Base Architecture & Llama-3.2-1B & Llama-3.2-1B & ResNet family \\
Number of Layers/Blocks & 16 & 16 & 16 (Blocks) \\
Hidden Dimension ($d_{model}$) & 2048 & 2048 & -- \\
FFN Inner Dimension ($d_{ffn}$) & 8192 & 8192 & -- \\
Number of Attention Heads & 32 & 32 & -- \\
Vocabulary Size & 50,280 & 50,280 & -- \\
Block Size & 2048 & 2048 & -- \\
Classes & -- & -- & 1000 \\ \bottomrule
\end{tabular}
\caption{\textbf{Architectural details of the primary models used in our experiments.} Configurations for the three architectures, spanning language and vision (Transformer, LayerSkip, ResNet-50).}
\label{tab:model_configs}
\end{table}

%% file: tables/tab_tapered_configs.tex
\begin{table}[h!]
\centering
\begin{tabular}{@{}llcl@{}}
\toprule
\textbf{Variant Name} & \textbf{Tapering Target(s)} & \textbf{Intensity} & \textbf{Governing Parameters} \\ \midrule
CascadeFormer-A$_1$ & Attention Dimension & Low & $F_d = 4$ \\
CascadeFormer-A$_2$ & Attention Dimension & High & $F_d = 2$ \\
\addlinespace
CascadeFormer-F$_1$ & FFN Dimension & Low & $S_f = 128$ \\
CascadeFormer-F$_2$ & FFN Dimension & High & $S_f = 256$ \\
\addlinespace
CascadeFormer-C$_1$ & Attention \& FFN & Low & $F_d = 4, \ S_f = 128$ \\
CascadeFormer-C$_2$ & Attention \& FFN & High & $F_d = 2, \ S_f = 256$ \\ \bottomrule
\end{tabular}
\caption{\textbf{Hyperparameter configurations for the CascadeFormer variants.} The parameters $F_d$ and $S_f$ control the tapering intensity for the attention and FFN dimensions, respectively. Lower $F_d$ and higher $S_f$ values correspond to more aggressive capacity reduction.}
\label{tab:tapered_configs}
\end{table}

%% file: tables/tab_training_hyperparams.tex
\begin{table}[h!]
\centering
\begin{tabular}{@{}lcc@{}}
\toprule
\textbf{Hyperparameter} & \textbf{Transformer models} & \textbf{ResNet-50} \\
\cmidrule(lr){1-3}
Dataset & pretokenized-dolma & ImageNet-1k \\
AdamW $\beta_1, \beta_2$ & (0.9, 0.95) & (0.9, 0.999) \\
Weight Decay & 0.1 & 0.05 \\
Learning Rate Schedule & Cosine w/ Warmup & Cosine w/ Warmup \\
Peak Learning Rate & 2e-4 & 5e-4 \\
Warmup Steps & 2000 & 5 Epochs \\
Batch Size & 64 & 512 \\
Epochs & 1 & 30 \\
Gradient Clipping Norm & 1.0 & - \\
\bottomrule
\end{tabular}
\caption{\textbf{Training hyperparameters for all experiments.} This table provides the complete set of hyperparameters used for training the language and vision models, ensuring full reproducibility of our results.}
\label{tab:training_hyperparams}
\end{table}

%% file: appendix/02_additional_results.tex
\section{Additional Results and Analyses}
\label{appx:additional_results}
We give additional layer-importance profiles, layer-wise trend plots, and a representational-similarity analysis that support the main findings.

\subsection{Gradient Fan-in Derivations} \label{app:pathcounts}

\paragraph{Definition and Counting Rule.}
We define gradient fan-in as the number of downstream transformation edges aggregated at a layer's input, $x_l$. This first-order proxy, counting the identity path, each downstream block’s Jacobian branch, and the final head, captures the structural asymmetry driving GFA. It is a count of contributing channels, not a combinatorial path enumeration. For a standard $N$-block stack with one head, the fan-in at layer $l$ is:
\begin{equation}
\phi_l = (N - l) + 2, \quad \text{(functional blocks + identity path + head)}
\label{eq:fanin_standard}
\end{equation}

\paragraph{Fan-in Under Deep Supervision.}
Architectures with deep supervision introduce an auxiliary loss head $\mathcal{L}_k$ at the output of each block $k$, with the total loss being $\mathcal{L}_{\text{total}} = \mathcal{L}_N + \sum_{k=0}^{N-1} \alpha_k \mathcal{L}_k$. The gradient at the input to block $l$, $g'_l$, now aggregates signals from all downstream loss functions:
\begin{equation}
g'_l = \frac{\partial \mathcal{L}_{\text{total}}}{\partial x_l} = \underbrace{\frac{\partial \mathcal{L}_N}{\partial x_l}}_{\text{Original Hierarchy}} + \sum_{k=l}^{N-1} \alpha_k \underbrace{\frac{\partial \mathcal{L}_k}{\partial x_l}}_{\text{Auxiliary Hierarchies}}.
\label{eq:deep_supervision_gradient}
\end{equation}
Each auxiliary loss $\mathcal{L}_k$ (sourced at $x_{k+1}$) effectively creates a new residual sub-network for its gradient to traverse back to $x_l$. The fan-in from a single auxiliary loss at layer $k$ to layer $l$ is analogous to a standard network of length $(k+1-l)$ and is thus $(k-l)+1+1 = (k-l+2)$. Assuming for simplicity that it is attached at $x_k$, the fan-in we use below is $(k-l)+1$.
The total fan-in $\phi'_l$ is the sum of the fan-in from the original path and all new auxiliary paths originating from layers $l$ through $N-1$:
\begin{equation}
\phi'_l = \phi_l + \sum_{k=l}^{N-1} ((k-l)+1),
\end{equation}
where $\phi_l = (N-l)+2$ is the standard single-head fan-in from Eq.~\ref{eq:fanin_standard}. Substituting $\phi_l$ and letting $j=k-l$, the summation becomes:
\begin{equation}
\phi'_l = ((N-l)+2) + \sum_{j=0}^{N-1-l} (j+1).
\end{equation}

Solving this arithmetic series and combining terms yields:
\begin{equation}
\phi'_l = ((N-l)+2) + \frac{(N-l)(N-l+1)}{2}.
\label{eq:deep_supervision_fanin_solved_app}
\end{equation}
The leading $\tfrac{1}{2}(N-l)^2$ term shows that deep supervision transforms the linear fan-in disparity into a quadratic one, amplifying GFA.

\paragraph{Fan-in with Virtual Depth.}
Consider 8 physical blocks unrolled into $N_{\text{virt}}{=}17$ virtual positions with shared parameters:
L1--L4 (1$\times$), L5 (2$\times$), L6 (3$\times$), L7 (3$\times$), L8 (5$\times$).
Let $V_i$ be the block at virtual position $i$ ($i{=}1,\ldots,17$). The fan-in for a physical layer $L_p$ sums over its virtual instances:

\begin{equation}
\text{FanIn}(L_p) \;=\; \sum_{i:\,V_i=L_p} \big(N_{\text{virt}} - i + 2\big).
\label{eq:virtual_fanin_app}
\end{equation}

This yields the following Standard $\to$ Virtual counts:
L1: 9$\to$18, L2: 8$\to$17, L3: 7$\to$16, L4: 6$\to$15, L5: 5$\to$27, L6: 4$\to$33, L7: 3$\to$24, L8: 2$\to$20.

\subsection{Layer-wise Gradient and Importance Trends}
The following figures provide a detailed visualization of the data from Figure~\ref{fig:grad_importance_correlation}.

\begin{figure}[htbp]
\centering
\includegraphics[width=0.8\textwidth]{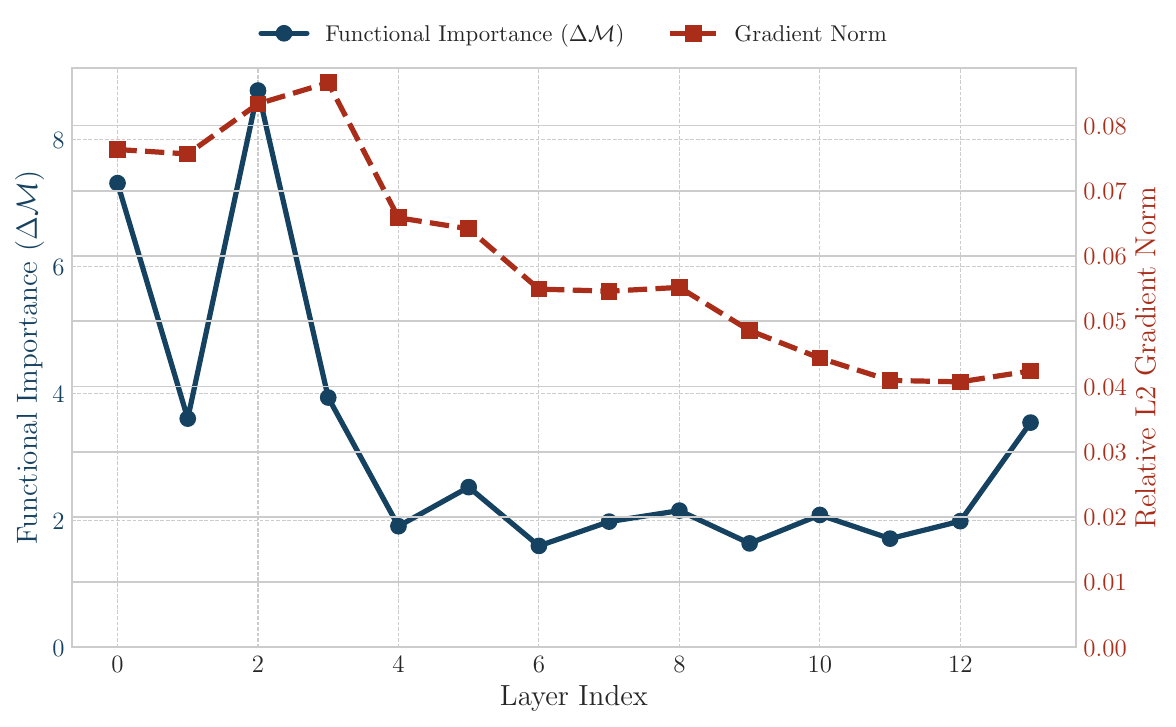}
\caption{\textbf{Layer-wise comparison for the Vanilla Transformer.} This figure details the relationship between relative L2 gradient norm (red, dashed) and functional importance (blue, solid). The correspondence visually reinforces the correlation ($\rho=0.62$) from Figure~\ref{fig:grad_importance_correlation}a.}
\label{fig:appendix_trends_vanilla}
\end{figure}

\begin{figure}[htbp]
\centering
\includegraphics[width=0.8\textwidth]{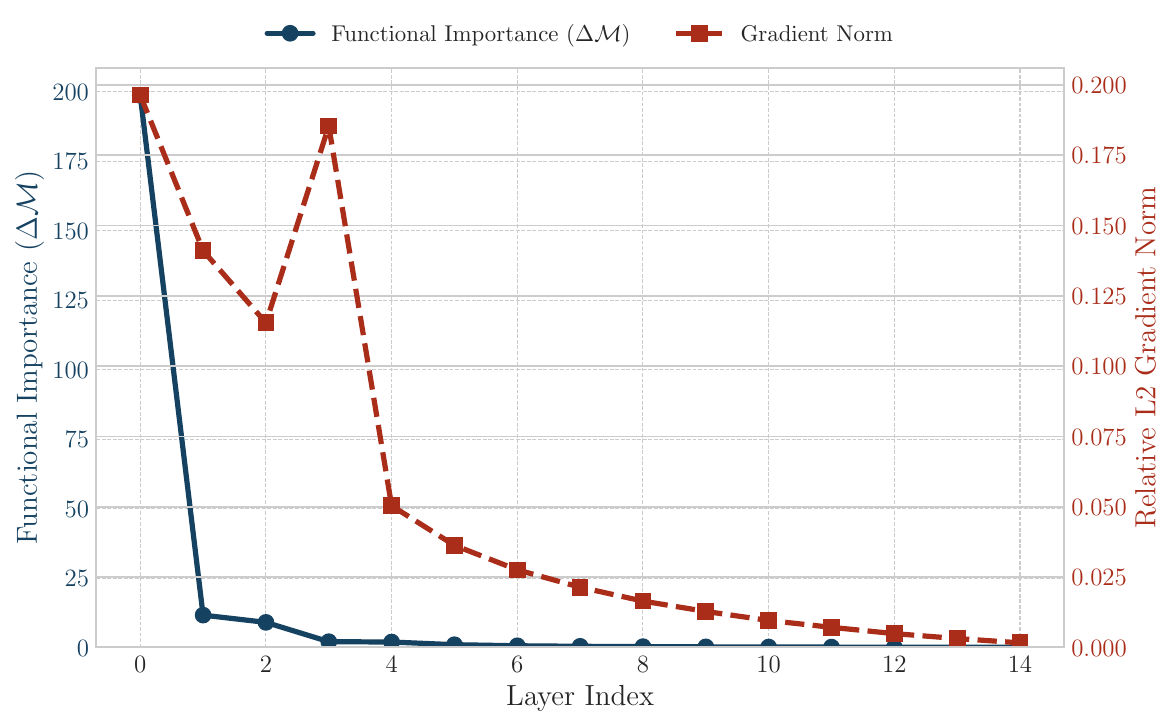}
\caption{\textbf{Layer-wise comparison for the LayerSkip Transformer.} This figure details the relationship between relative L2 gradient norm (red, dashed) and functional importance (blue, solid). The curves track closely, matching the near-perfect correlation ($\rho=0.99$) from Figure~\ref{fig:grad_importance_correlation}b.}
\label{fig:appendix_trends_layerskip}
\end{figure}

\begin{figure}[htbp]
\centering
\includegraphics[width=0.8\textwidth]{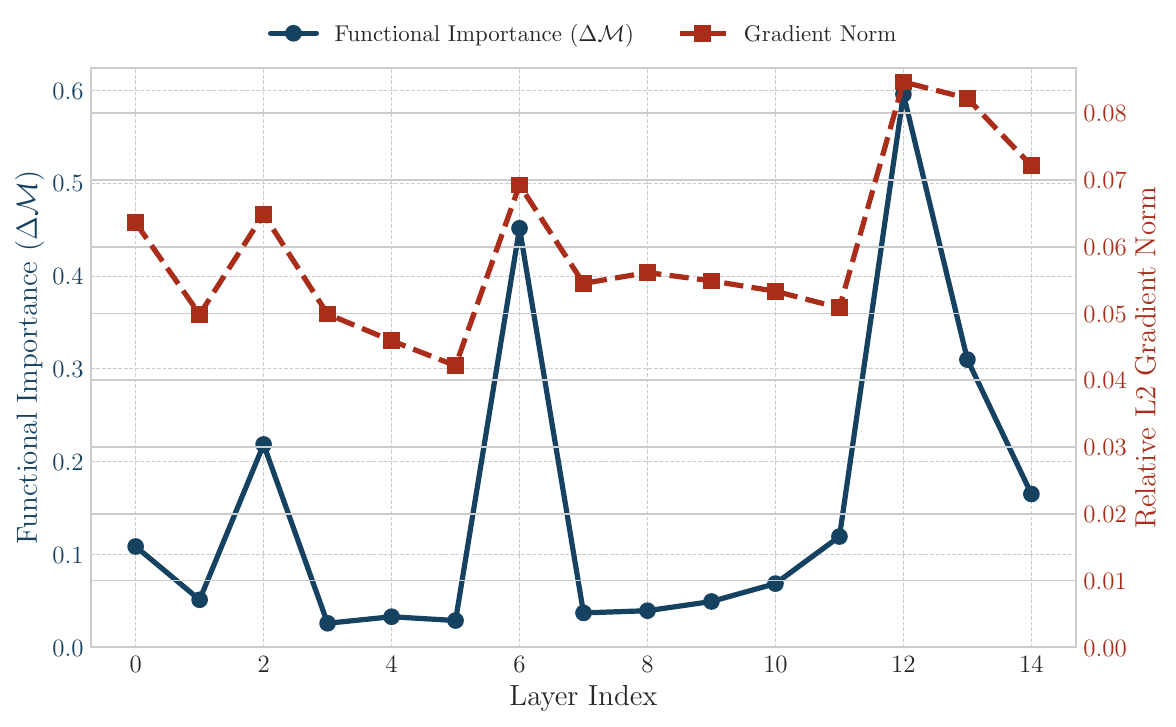}
\caption{\textbf{Layer-wise comparison for ResNet-50.} This figure details the relationship between relative L2 gradient norm (red, dashed) and functional importance (blue, solid). A clear positive relationship is evident, corroborating the correlation ($\rho=0.83$) from Figure~\ref{fig:grad_importance_correlation}c.}
\label{fig:appendix_trends_resnet}
\end{figure}